\documentclass{bmvc2k}
\usepackage{float}
\usepackage{booktabs}
\usepackage{multirow}
\usepackage{adjustbox}
\usepackage{transparent}
\usepackage[hypcap=false]{caption}
\usepackage[accsupp]{axessibility}  

\definecolor{rebuttalcolor}{RGB}{0,165,135}

\title{NCA-Morph: Medical Image Registration with Neural Cellular Automata}

\addauthor{Amin Ranem}{amin.ranem@gris.informatik.tu-darmstadt.de}{1}
\addauthor{John Kalkhof \\ Anirban Mukhopadhyay}{}{1}

\addinstitution{
 Technical University of Darmstadt,\\
 Karolinenpl. 5,\\
 64289 Darmstadt, Germany
}

\def\etal{\emph{et al}\bmvaOneDot}

\runninghead{Ranem \etal}{NCA-Morph}

\begin{document}

\maketitle

\begin{abstract}
Medical image registration is a critical process that aligns various patient scans, facilitating tasks like diagnosis, surgical planning, and tracking. Traditional optimization-based methods are slow, prompting the use of Deep Learning (DL) techniques, such as VoxelMorph and Transformer-based strategies, for faster results. However, these DL methods often impose significant resource demands.
In response to these challenges, we present \textbf{NCA-Morph}, an innovative approach that seamlessly blends DL with a bio-inspired communication and networking approach, enabled by Neural Cellular Automata (NCAs). NCA-Morph not only harnesses the power of DL for efficient image registration but also builds a network of local communications between cells and respective voxels over time, mimicking the interaction observed in living systems.
In our extensive experiments, we subject NCA-Morph to evaluations across three distinct 3D registration tasks, encompassing Brain, Prostate and Hippocampus images from both healthy and diseased patients. The results showcase NCA-Morph's ability to achieve state-of-the-art performance. Notably, NCA-Morph distinguishes itself as a lightweight architecture with significantly fewer parameters; 60\% and 99.7\% less than VoxelMorph and TransMorph. This characteristic positions NCA-Morph as an ideal solution for resource-constrained medical applications, such as primary care settings and operating rooms.
\end{abstract}


\section{Introduction}
\label{sec:intro}


In healthcare, the process of image registration is a core technical problem to combine different images of one (or more) patients \cite{hill2001medical, zitova2003image} into a common spatio-temporal coordinate system \cite{brown1992survey}. Successful registration enables a wide range of applications from diagnosis to surgical planning and tracking \cite{eggers2006image, ji2014cortical, maintz1998survey, uneri20133d}. The registration process infers the deformation of a medical image by using a \textit{displacement vector field} \cite{nagel1986investigation} or \textit{deformation field} \cite{meijering1999image}. 
Traditional -- non-learning -- registration is a time-consuming process that requires optimizing the deformation field based on a set of parameters \cite{avants2008symmetric, bajcsy1989multiresolution, thirion1998image}. The introduction of Deep Learning (DL) methods like VoxelMorph- and Transformer-based strategies \cite{balakrishnan2019voxelmorph, chen2021vit, perez2023learning} allow learning the deformation field end-to-end without requiring a ground truth warp field. Though DL makes inference faster, the computational requirement is ever increasing. In scenarios like cluttered operating rooms or primary care disease progression analysis, where deploying large computers is often impractical, the need for low-resource inference becomes crucial for democratizing medical Artificial Intelligence \cite{fuchs2023closing}.

\begin{figure}[ht!]
\begin{center}
\includegraphics[trim=0 7cm 0 0, clip, width=\linewidth]{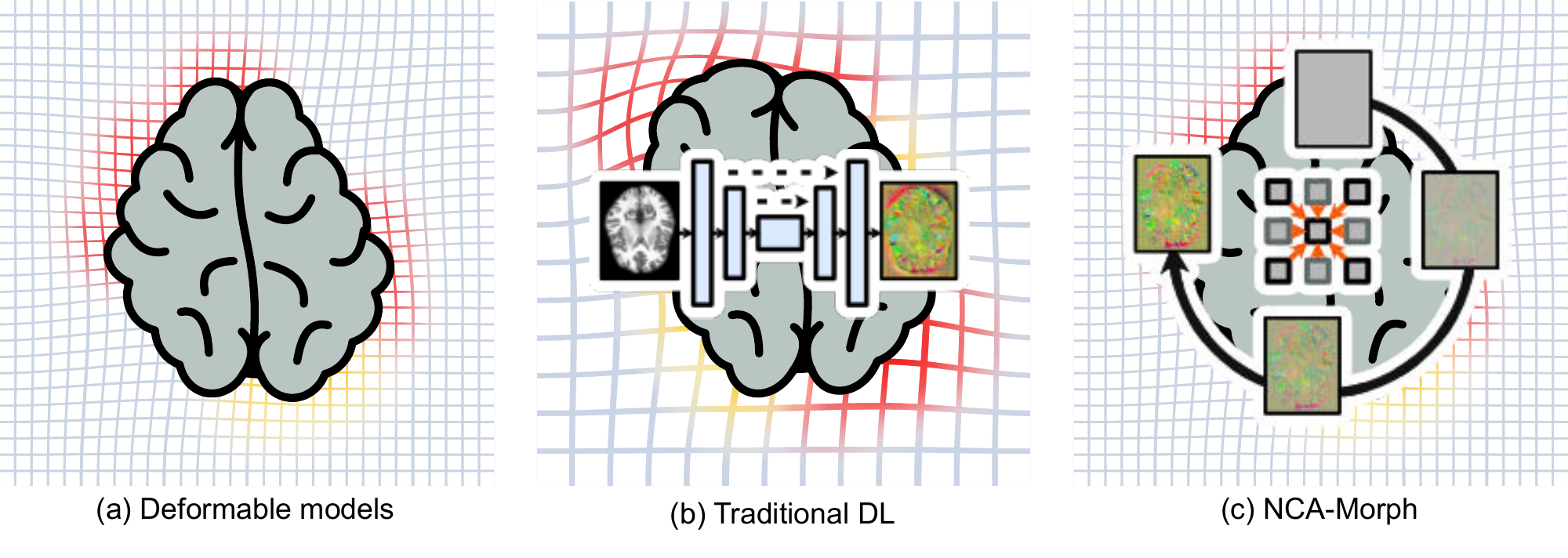}
\end{center}
\caption{(a) Traditional optimization techniques like SyN focus on \textit{local displacements of voxels} while optimizing a set of parameters during inference \cite{beg2005computing, thirion1998image}. (b) Traditional DL techniques leverage global information while estimating a \textit{one-shot deformation field during inference} based on displacement rules learned during training \cite{balakrishnan2019voxelmorph, chen2022transmorph, kim2021cyclemorph}. (c) NCA-Morph learns \textit{local displacement rules} in a DL-fashion while \textit{regularizing the deformation field during inference} based on local interactions and learned rules.}
\label{fig:story_fig}
\end{figure}

In this paper, we introduce a bio-inspired paradigm for image registration as an emergent phenomenon, drawing inspiration from the dynamics of cellular communication and collective intelligence, Figure \ref{fig:story_fig}. 
Building on this, we explore the recent advances in Neural Cellular Automata (NCA) for lightweight and adaptive medical applications. NCA is an emerging field with promise for various applications where a common local rule is uniformly applied to all cells in an image or 3D volume. The particular appeal of NCAs lies in their ability to communicate solely with their nearby neighbors. A complex goal can be reached by repeating this same rule multiple times. The one-cell architecture makes NCAs lightweight with only around 13k parameters. To date, most publications in this area have been limited to toy examples, such as growing an image from a single cell \cite{mordvintsev2020growing}, self-classifying robots \cite{walker2022physical}, and segmentation \cite{sandler2020image} on image sizes of 64x64. This is partially due to the substantial video random access memory (VRAM) requirements during training and the more difficult convergence in general. However, recent developments like M3D-NCA \cite{kalkhof2023med} have expanded the potential use cases.

Given the success of NCAs in segmenting medical images \cite{kalkhof2023m3d}, the question naturally arises as to whether they can be applied to another important area of medical image analysis: \textit{registration}. However, this problem poses a greater challenge since the goal of image registration is generally to match two 3D volumes.
Our NCA-based approach stands apart, as it does not involve variational minimization, elastic deformations or convolutional networks and spatial transformations like traditional and DL approaches do. It offers a novel, efficient pathway for achieving high-precision registration by leveraging the self-organizing properties of NCAs.

NCAs are initially conceived as a \textit{solitary active cell} that can be seen as mimicking a single voxel taking the initiative to build a network of communication between its neighboring cells. These connections facilitate communication and collaboration among cells to collectively influence their transformations, i.e. a 3D deformation field for optimal registration.
During the training of the NCA, the initial network of connections between the cells evolves and matures into a resilient and complex communication network among the NCA cells. Remarkably, despite the growing complexity of the network, each individual cell adheres to the same rule, mirroring the uniformity of action within a developing organism.

We demonstrate how this bio-inspired approach is (1) \textbf{parameter efficient}, (2) holds the potential to significantly \textbf{enhance the precision and efficiency of medical image registration} while (3) \textbf{being light-weight} and able to run on a Raspberry Pi. 


\textbf{NCA-Morph} establishes the basis for utilizing NCA architectures in the context of medical image registration. The contributions of this work are two-fold: (1) NCA-Morph successfully introduces the \textit{NCAs in 3D medical image registration} by learning local displacement rules. (2) We perform an extensive \textit{analysis of the NCA architecture and its components} on three different organs (Brains, Prostate and Hippocampus) in the form of inter-subject setups to get a proper understanding of their impact on medical image registration. 
The results show that our method achieves accuracy comparable to the state-of-the-art. With the help of \emph{NCA-Morph}, registration can be completed in a few seconds using a GPU while being a lightweight architecture with $60\%$ and $99.7\%$ less parameters compared to VoxelMorph and TransMorph respectively.


\section{Related Work}
\label{app:rel_work}

\paragraph{Traditional medical image registration:}
There are several traditional 3D medical image registration techniques based on basic -- non-learning -- assumptions like the concept of \textit{attraction} using a deformable model like Maxwell’s demons \cite{thirion1998image}. Another strategy uses large deformation metric mappings via geodesic flows \cite{beg2005computing}, which is based on variational minimization on vector fields using Euler-Lagrange equations. Topology-preserving diffeomorphic transformations have demonstrated success in diverse studies related to computational anatomy. Approaches like elastic matching \cite{bajcsy1989multiresolution} or symmetric diffeomorphic registration \cite{avants2008symmetric} are based on local elastic deformations in combination with methods to maximize the cross-correlation for diffeomorphic cases in the space of topology preserving maps. Different work uses a Levenberg–Marquardt strategy and a constant Eulerian velocity framework \cite{ashburner2007fast} to perform registration. One of the well-known traditional registration strategies is standard symmetric normalization, SyN and SyNCC using Normalized Cross Correlation \cite{avants2008symmetric}. However, traditional methods like SyNCC are known for their slow processing speed.

\paragraph{(Deep-)Learning-based medical image registration:}
Recent research focuses on DL-based techniques for medical image registration. For instance, there are methods that are trained end-to-end using segmentations as their ground truth or optical flows like warp fields \cite{krebs2017robust, rohe2017svf, sokooti2017nonrigid, yang2017quicksilver}. Being dependent on segmentations or ground truth warp fields for training is a crucial bottleneck as to why another group of methods has emerged that focuses on unsupervised medical image registration.

The prime example of unsupervised medical image registration is the VoxelMorph framework \cite{balakrishnan2019voxelmorph}. VoxelMorph and several widely adopted methods \cite{cao2017deformable, de2017end, li2017non, meng2023non} rely on Convolutional Neural Network (CNN) structures along with a spatial transformation function \cite{jaderberg2015spatial}. This function is employed to deform images by applying an estimated deformation field denoted as $\phi$.

Since the publication of VoxelMorph, advanced and more complex methods based on this framework were published. Notably, given the current success of Transformers in the field of medical imaging, recent publications have explored image registration using Transformer- or Attention-based networks \cite{chen2022transmorph, chen2021vit, chen2023deformable, perez2023learning, li2022dual, shi2022xmorpher}. Additionally, techniques like CycleMorph incorporate cycle loss and local cross-correlation during their training process \cite{kim2021cyclemorph}.

In contrast, our NCA-Morph introduces a bio-inspired method for image registration, following the adaptability observed in living organisms. NCA-Morph enhances the precision and adaptability of medical image alignment by guiding the gradual evolution of the cells in the NCA network. This approach sets itself apart by achieving superior performance with notable parameter efficiency, offering a lightweight solution for precise medical image registration.

\section{Methodology}
To our knowledge, no work to date addresses the problem of 3D registration using NCAs.

Let $\mathcal{O} \subset \Omega$ be a dataset $\mathcal{O}$, while $\Omega \subset \mathbb{R}^3$ is a 3D spatial domain. $\mathcal{O}$ consists of $m$ pairs $\left\{\left(I^k_{img}, I^k_{seg}\right)\right\}_{k \leq \lvert \mathcal{O} \rvert}$, where $I_{img}$ is a 3D scan and $I_{seg}$ the corresponding segmentation mask. For medical image registration, patient pairs are used, whereas one image pair $k$ is referenced as \textbf{fixed (F)} and the other one as \textbf{moving (M)} -- $\left\{\left(F^k_{img}, F^k_{seg}\right) \mid \left(M^k_{img}, M^k_{seg}\right)\right\}$. Registration is a Transformation $\mathcal{T}$ which maps an image pair from the \textit{moving}, i.e. \textit{source} domain $\Omega_{M}$ into the \textit{target} domain $\Omega_{F}$ of the fixed image pair using a parameteric representation $\phi$ of the deformation field: $\mathcal{T}_{\phi}\left(M^k_{img}, M^k_{seg}\right) \cong \phi \circ \left[\left(M^k_{img}, M^k_{seg}\right) \in \Omega_{M}\right] = \left(M_{\phi(img)}^k, M_{\phi(seg)}^k\right) \in \Omega_{F}.$


\begin{figure}[ht!]
    \begin{center}
    \includegraphics[width=0.9\textwidth]{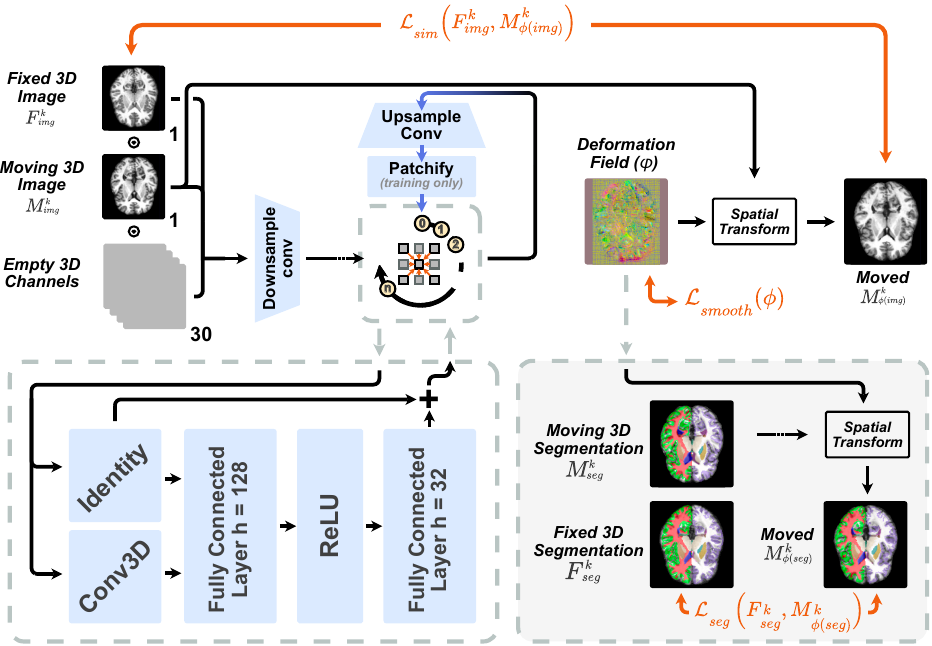}
    \end{center}
    \caption{\textit{NCA-Morph architecture for registration}: The input is a concatenation $\odot$ of the fixed and moving image accompanied by 30 empty channels. Upon reducing the image size by a quarter, the initial NCA forecasts the overall flow of cells. Subsequently, this rough flow is combined with the detailed images, and a second NCA completes the deformation. While the second NCA solely observes patches during training, it operates at full resolution during inference. The resulting deformation field can be applied to warp both the moving image and its segmentation.}
    \label{fig:nca_arch}
\end{figure}

\paragraph{Neural Cellular Automata for registration:}
Different NCA variations are introduced throughout this work based on their settings using the NCA-Morph$^{\text{steps}}_{\text{kernel}}$ notation. When designing \textit{NCA-Morph} for registration, we formulate a localized update rule represented as $g_{\theta}(F, M) = u$. This rule acts to incrementally adjust the cell state of each voxel, progressively aligning it with a deformation field $\phi$. Through a series of iterations we gradually approach the desired deformation field by applying this rule locally to every voxel within a 3D image.

We define $\phi$ as the accumulated outcome of applying the update function $g$ over a series of time steps. This can be expressed as $\phi = g(x_{t=n}) \odot g(x_{t=n-1}) \odot \cdots \odot g(x_{t=0})$, where $\odot$ represents a simple concatenation, $t$ are the time steps and $x$ represents the combined 3D volume of the moving image and the fixed image: $x_0 := F^k_{img} \odot M^k_{img}$. The objective is to identify anatomically similar regions in these images.

In Figure \ref{fig:nca_arch}, we give an overview of our NCA-Morph architecture and the underlying inference setup. We initiate the model with the input images $M^k_{img}$ and $F^k_{img}$. At each time step $t$, we execute the function $g$, which is parameterized by $\theta$. This function guides updates for each cell in the input images, gradually refining the alignment, i.e. \textit{propagation of the deformation field over time} following our bio-inspired communication approach among cells. We train these parameters by applying $\phi$ alongside a spatial transformation function, resulting in $M_{\phi(img)}^k$. This transformation enables the computation of the similarity loss $\mathcal{L}_{sim}\left(F^k_{img}, M_{\phi(img)}^k\right)$ between the \textit{fixed} and \textit{moved} images, facilitating the alignment process.

Additionally, we incorporate a smoothing technique $\mathcal{L}_{smooth}(\phi)$ for the deformation field, as proposed in VoxelMorph, enforcing a spatially smooth deformation. Since we possess segmentation masks, we include a segmentation loss term $\mathcal{L}_{seg}\left(F^k_{seg}, M_{\phi(seg)}^k\right)$, measuring the alignment between the fixed and moved segmentation masks.


\textbf{Model architecture:} 
M3D-NCA \cite{kalkhof2023m3d}, initially designed for medical segmentation, presents limitations when applied to registration tasks. This is because registration requires the alignment of two volumes, which demands a different architecture to handle the increased complexity. Our introduced NCA-Morph uses an n-level approach, where NCAs are running on different scales of the image, starting from a coarse to a fine worldview, increasing the global information. 
This global information is then upscaled to the next image scale and can thus be combined with high-resolution image information. 
NCA-Morph solves the high VRAM requirements by using patches on the high-resolution layers of the model. Due to the one-cell architecture, NCAs can be trained on patches and inferred on the full-scale image.

The update rule on each level $g$ is represented by an NCA, which uses one 3D convolution with a kernel size of $k=3$ as its input, describing a simplification of all neighborhood information into a vector $v_n$. We concatenate this information $\left(\odot\right)$ with the cell's current state $c$, which leaves us with a state vector of $v = v_n \odot c$. In the next step, $v$ is updated by two fully connected hidden layers, where the first one maps $v$ onto a vector of size $h=128$, the Rectified Linear Unit (ReLU) activation function is applied and the second layer maps it back to the input channel size $i_c$, resulting in a vector $v_i$. We then update $c$ by adding $v_i$ thus making $c_{t+1} = c_{t} + v_i$.



\section{Datasets}
\label{sec:data}
\paragraph{Open Access Series of Imaging Studies (OASIS):}
The OASIS-1 dataset \cite{marcus2007open} is a cross-sectional collection of 416 subjects, aged from 18 to 96. For each subject, up to four individual T1-weighted MRI scans were obtained during a single session. 
The dataset was obtained within the \textit{Learn2Reg} Challenge \cite{hering2022learn2reg}. For this work, we randomly sampled 364 inter-patient image pairs from the training distribution and another 50 inter-patient pairs from the testing distribution.

\paragraph{Prostate task:}
Our prostate registration focuses on two distinct datasets: the I2CVB
\cite{lemaitre2015computer} and PROMISE12
dataset \cite{litjens2014evaluation}. These datasets are used in their original form as part of the \emph{Multi-site Dataset for Prostate MRI Segmentation Challenge} \cite{liu2020saml,liu2020ms,down_side}. They consist of T2-weighted MRIs. For our registration preparation, we randomly selected 240 image pairs between different patients for training and an additional 60 pairs for evaluation.

\paragraph{Hippocampus task:}
The registration task for the Hippocampus involves two datasets: the \emph{Harmonized Hippocampal Protocol} dataset \cite{boccardi2015training} and the \emph{Dryad} dataset \cite{kulaga2015multi}. The Hippocampus task comprises T1-weighted MRIs of elderly healthy individuals and patients with Alzheimer’s disease. We follow the same registration preparation as for prostate (240-60).

\section{Results}
We conduct a comprehensive analysis on computation vs. performance of our NCA-Morph variations in comparison to traditional methods like SyN \cite{avants2008symmetric} and state-of-the-art registration techniques, such as VoxelMorph \cite{balakrishnan2019voxelmorph}, TransMorph \cite{chen2022transmorph}, ViTVNet \cite{chen2021vit}, and NICE-Trans \cite{meng2023non}.

\subsection{Computation vs. Performance}
\label{sec:res}
\begin{figure}[htb]
\begin{center}
\includegraphics[width=\linewidth]{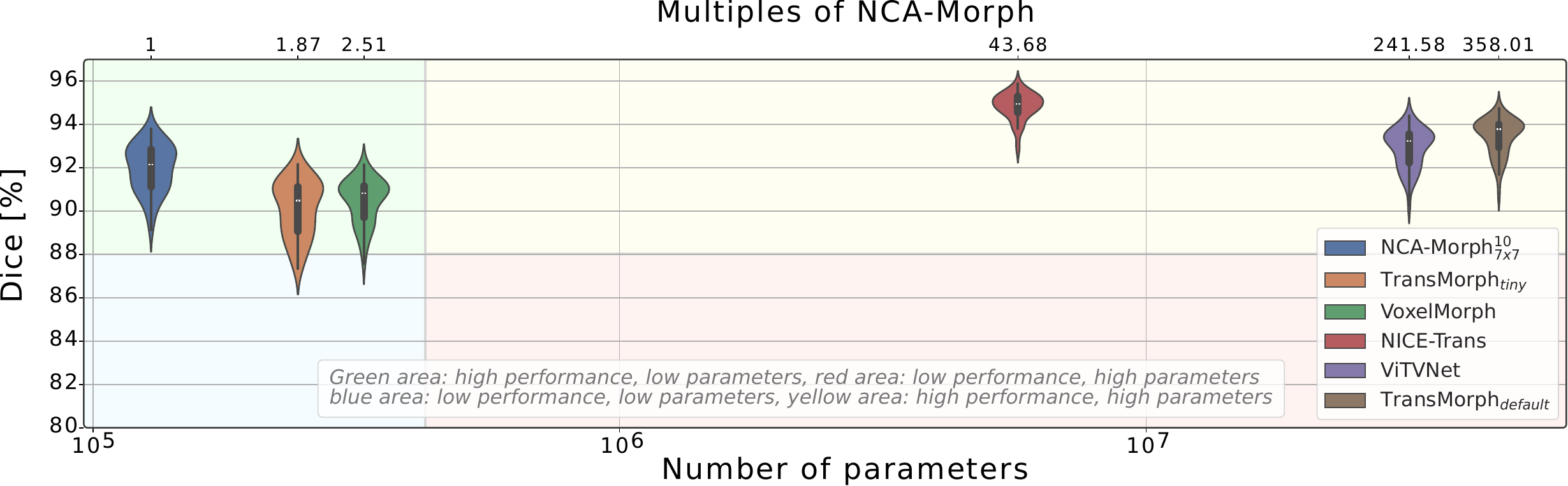} 
\end{center}
\caption{NCA-Morph compared to VoxelMorph, TransMorph, ViTVNet and NICE-Trans in terms of the number of parameters and Dice performance based on the OASIS registration task. NCA-Morph uses 60\% and 99.7\% fewer parameters than VoxelMorph and TransMorph.}
\label{fig:comp_res}
\end{figure}

To get a proper comparison over the computational burden of our proposed NCA-Morph variations and state-of-the-art (SOTA) methods, the Dice performance distribution across the number of model parameters is shown in Figure \ref{fig:comp_res} and Table \ref{tab:res}.

Figure \ref{fig:comp_res} shows that the NCA-Morph has a significant advantage in terms of the number of parameters. Our best NCA-Morph achieves a superior Dice performance than the SOTA VoxelMorph while using $60\%$ less parameters.

Table \ref{tab:res} gives a proper overview of allocated resources during training and inference. Jacobian  determinants are used as a measure for deformation fields as negative determinants indicate that the transformation causes local volume compression or folding. The runtime indicates the number of seconds to train a network for a single epoch along with the model size in megabyte (MB). The number of parameters describes the amount of trainable model parameters, whereas the inference time to register a single 3D volume on a GPU is indicated as "GPU sec". 
Although our NCA-Morph takes 3 seconds more for a single registration, it has a significantly lower number of parameters compared to VoxelMorph and TransMorph, while achieving still similar performances.


\begin{table}[h]
\begin{center}
\begin{adjustbox}{max width=\linewidth}
{\begin{tabular}{clccc|cccc}\toprule
  Anatomy & Method & SSIM $\uparrow$ & $\left\lvert J_\phi \right\rvert \leq 0 \downarrow$ & Dice $\uparrow$ [\%] & Runtime $\downarrow$ [sec] & Model size $\downarrow$ [MB] & Nr. parameters $\downarrow$ & GPU sec $\downarrow$ \\ \hline \hline
    \multirow{11}{*}{\rotatebox{90}{Brain (OASIS)}} 
    & SyN & $74.69 \; (1.26)$ & $--$ & $68.58 \; (1.56)$ & -- & -- & -- & 11.38$^{\rtimes}$ \\
    & SyNCC & $78.93 \; (1.33)$ & $--$ & $72.00 \; (1.53)$ & -- & -- & -- & 448$^{\rtimes}$ \\
    \cline{2-9}
    & VoxelMorph & $82.84 \; (1.36)$ & $\mathbf{254217 \; (20553)}$ & $90.82 \; (1.03)$ & 722 & 1.30 & 327,331 & 7.79 \\
    & TransMorph$_{\text{tiny}}$ & $81.20 \; (1.53)$ & $377100 \; (32541)$ & $90.49 \; (1.29)$ & 361 & 4.60 & 244,527 & 6.62 \\
    & TransMorph$_{\text{default}}$ & $86.21 \; (1.26)$ & $283837 \; (20883)$ & $93.80 \; (0.82)$ & 445 & 190.70 & 46,771,251 & 10.89 \\
    & ViTVNet & $84.76 \; (1.38)$ & $310839 \; (25593)$ & $93.25 \; (0.88)$ & 551 & 126.3 & 31,560,079 & 8.98 \\
    & NICE-Trans & $\mathbf{88.31 \; (1.29)}$ & $305220 \; (29010)$ & $\mathbf{94.95 \; (0.62)}$ & 654 & 24.90 & 5,706,508 & 6.59 \\
    \cline{2-9}
    & NCA$^{5}_{3\times3}$ & $81.00 \; (1.70)$ & $291620 \; (21842)$ & $90.99 \; (1.20)$ & 560 & 0.52 & 129,504 & 11.51 \\
    & NCA$*^{5}_{3\times3}$ & $65.45 \; (2.36)$ & $323441 \; (25429)$ & $90.67 \; (1.22)$ & 567 & 0.52 & 130,641 & 9.90 \\
    & NCA$^{10}_{7\times7}$ & $\mathbf{82.70 \; (1.66)}$ & $\mathbf{257241 \; (24826)}$ & $\mathbf{92.14 \; (1.09)}$ & 630 & 0.52 & 129,504 & 16.79 \\
    & NCA$*^{10}_{7\times7}$ & $82.05 \; (1.57)$ & $336983 \; (23095)$ & $91.29 \; (1.18)$ & 636 & 0.53 & 130,641 & 12.92 \\
    \hline \hline
    \multirow{11}{*}{\rotatebox{90}{\shortstack{Prostate}}}
    & SyN & $95.50 \; (1.09)$ & $--$ & $49.32 \; (21.02)$ & -- & -- & -- & 7.22$^{\rtimes}$ \\
    & SyNCC & $95.59 \; (1.15)$ & $--$ & $51.88 \; (22.13)$ & -- & -- & -- & 420$^{\rtimes}$ \\
    \cline{2-9}
    & VoxelMorph$^{\circledast}$ & $\mathbf{95.24 \; (1.31)
}$ & $\mathbf{0.00 \; (0.00)}$ & $\mathbf{53.39 \; (16.11)}$ & 282 & 1.30 & 327,331 & 7.44 \\
    & TransMorph$_{\text{tiny}}$ & $95.24 \; (1.31)$ & $0.00 \; (0.00)$ & $53.38 \; (16.16)$ & 294 & 4.60 & 244,527 & 6.49 \\
    & TransMorph$_{\text{default}}$ & $95.23 \; (1.22)$ & $96881 \; (94986)$ & $51.78 \; (15.82)$ & 367 & 190.70 & 46,771,251 & 6.68 \\
    & ViTVNet & $94.98 \; (1.04)$ & $111511  \; (72308)$ & $27.61 \; (19.76)$ & 417 & 126.30 & 31,560,079 & 6.34 \\
    & NICE-Trans$^{\circleddash}$ & $96.06 \; (0.91)$ & $2343699 \; (26870)$ & $0.00 \; (0.00)$ & 520 & 24.90 & 5,706,508 & 7.44 \\
    \cline{2-9}
    & NCA$^{5}_{3\times3}$ & $\mathbf{95.19 \; (1.31)}$ & $\mathbf{53195 \; (7134)}$ & $\mathbf{53.17 \; (15.80)}$ & 428 & 0.52 & 129,504 & 9.43 \\
    & NCA$*^{5}_{3\times3}$ & $95.12 \; (1.13)$ & $107357 \; (19479)$ & $35.53 \; (18.44)$ & 430 & 0.53 & 130,641 & 8.09 \\
    & NCA$^{10}_{7\times7}$ & $94.95 \; (1.30)$ & $333012 \; (29795)$ & $46.18 \; (15.54)$ & 477 & 0.52 & 129,504 & 13.30 \\
    & NCA$*^{10}_{7\times7}$ & $95.13 \; (1.10)$ & $83298 \; (14387)$ & $43.70 \; (15.46)$ & 479 & 0.53 & 130,641 & 10.25 \\
    \hline \hline
    \multirow{11}{*}{\rotatebox{90}{\shortstack{Hippocampus}}}
    & SyN & $83.49 \; (4.19)$ & $--$ & $35.26 \; (24.61)$ & -- & -- & -- & 0.35$^{\rtimes}$ \\
    & SyNCC & $83.24 \; (4.91)$ & $--$ & $32.03 \; (27.12)$ & -- & -- & -- & 14.21$^{\rtimes}$ \\
    \cline{2-9}
    & VoxelMorph$^{\circledast}$ & $80.76 \; (3.46)$ & $\mathbf{6.36 \; (48.9)}$ & $32.4 \; (22.75)$ & 33 & 1.30 & 327,331 & 0.68 \\
    & TransMorph$_{\text{tiny}}$ & $79.67 \; (4.47)$ & $49379 \; (8187)$ & $33.64 \; (25.97)$ & 62 & 4.60 & 244,527 & 0.56 \\
    & TransMorph$_{\text{default}}$ & $\mathbf{84.70 \; (4.52)}$ & $58104 \; (9120)$ & $\mathbf{48.28 \; (33.41)}$ & 69 & 190.70 & 46,771,251 & 1.08 \\
    & ViTVNet & $82.23 \; (3.83)$ & $29941 \; (7983)$ & $33.64 \; (26.71)$ & 66 & 126.30 & 31,560,079 & 0.93 \\
    & NICE-Trans$^{\circledcirc}$ & $67.82 \; (6.08)$ & $104669 \; (5135)$ & $4.80 \; (5.09)$ & 96 & 24.90 & 5,706,508 & 1.33 \\
    \cline{2-9}
    & NCA$^{5}_{3\times3}$ & $82.08 \; (3.92)$ & $29393 \; (8424)$ & $36.05 \; (27.88)$ & 45 & 0.52 & 129,504 & 0.53 \\
    & NCA$*^{5}_{3\times3}$ & $80.77 \; (4.25)$ & $44966 \; (11952)$ & $35.46 \; (27.34)$ & 42 & 0.53 & 130,641 & 0.57 \\
    & NCA$^{10}_{7\times7}$ & $\mathbf{82.85 \; (3.81)}$ & $\mathbf{28982\; (8357)}$ & $\mathbf{37.77 \; (28.54)}$ & 57 & 0.52 & 129,504 & 0.46 \\
    & NCA$*^{10}_{7\times7}$ & $80.80 \; (4.34)$ & $51756 \; (14247)$ & $33.18 \; (27.39)$ & 55 & 0.53 & 130,641 & 0.46 \\
    \bottomrule
\end{tabular}}
\end{adjustbox}
\end{center}
\caption{\textit{Computation vs. Performance}: Structural Similarity Index Measure (SSIM), negative Jacobian and Dice score for every trained DL model with runtime per epoch, model size, number of parameters and inference time on a GPU. For traditional methods, inference time on CPU is provided and marked with $^{\rtimes}$. NCA$*$ generates the deformation field with an additional CNN layer. Due to non-convergence: $\circledast$ is trained for 150 epochs, $\circledcirc$ for 100 epochs and $\circleddash$ performance dropped to 0\% after few epochs of training.}
\label{tab:res}
\end{table}

The outcomes and resource allocations for our most successful NCA-Morph ablation – 7x7 kernel, 10 steps, 32 channels, 128 hidden size and directly extracting the flow from the NCA-Morph – in addition to VoxelMorph, two TransMorph ablations, the ViTVNet and NICE-Trans distinctly illustrate that NCA-Morph models use significantly fewer parameters than all other models. Compared to the default TransMorph model, our \textit{NCA-Morph uses 99.7\% less parameters while achieving nearly identical Dice performance.} 

\subsection{NCA-Morph ablation studies}
\label{app:ablation}
We conducted a series of ablation experiments to investigate the impact of various hyperparameters on the performance of our NCA-Morph. Specifically, we explored the effects of altering the kernel size, the number of steps, channel size, and hidden size. Our objective was to understand which of these alterations lead to substantial changes in model performance to correctly determine a well suited setup for registration, Table \ref{tab:ablations}.

\begin{table}[ht]%
\begin{center}
\begin{adjustbox}{max width=0.7\linewidth}
    {\begin{tabular}{lccc}\toprule
      Method & SSIM $\uparrow$ & $\left\lvert J_\phi \right\rvert \leq 0 \downarrow$& Dice $\uparrow$ [\%] \\ \hline \hline
      \multicolumn{4}{c}{\textit{\textbf{Kernel size ablation}  (10 steps, 16 channels, 64 hidden size)}} \\
      \hline
        NCA$^{10}_{\mathbf{3}\times\mathbf{3}}$ & $\mathbf{78.82 \; (1.44)}$ & $456804 \; (22491)$ & $\mathbf{88.99 \; (1.44)}$ \\
        NCA$^{10}_{\mathbf{5}\times\mathbf{5}}$ & $78.83 \; (1.48)$ & $460264 \; (22363)$ & $88.95 \; (1.41)$ \\
        NCA$^{10}_{\mathbf{7}\times\mathbf{7}}$ & $78.93 \; (1.53)$ & $445223 \; (23424)$ & $88.91 \; (1.43)$ \\
        NCA$^{10}_{\mathbf{9}\times\mathbf{9}}$ & $78.88 \; (1.49)$ & $\mathbf{444250 \; (22278)}$ & $88.96 \; (1.43)$ \\
        \hline
      \multicolumn{4}{c}{\textit{\textbf{Step size ablation} ($3\times3$ kernel, 16 channels, 64 hidden size)}} \\
      \hline
        NCA$^{\mathbf{5}}_{3\times3}$ & $79.14 (1.49)$ & $\mathbf{360049 \; (27372)}$ & $\mathbf{89.13 \; (1.27)}$ \\
        NCA$^{\mathbf{10}}_{3\times3}$ & $78.82 \; (1.44)$ & $456804 \; (22491)$ & $88.99 \; (1.44)$ \\
        NCA$^{\mathbf{30}}_{3\times3}$ & $78.47 \; (1.46)$ & $445687 \; (20268)$ & $88.96 \; (1.42)$ \\
        NCA$^{\mathbf{50}}_{3\times3}$ & $\mathbf{78.45 \; (1.47)}$ & $436240 \; (20486)$ & $88.59 \; (1.53)$ \\
        \hline
      \multicolumn{4}{c}{\textit{\textbf{Channel and hidden size ablation} ($3\times3$ kernel, 5 steps)}} \\
      \hline
        NCA$^{5}_{3\times3}(16, 64)$ & $79.14 (1.49)$ & $360049 \; (27372)$ & $89.13 \; (1.27)$ \\
        NCA$^{5}_{3\times3}(16, 128)$ & $78.64 \; (1.41)$ & $439126 \; (24152)$ & $88.22 \; (1.40)$ \\
        NCA$^{5}_{3\times3}(32, 64)$ & $80.68 \; (1.64)$ & $315900 \; (30167)$ & $90.57 \; (1.21)$ \\
        NCA$^{5}_{3\times3}(32, 128)$ & $\mathbf{81.00 \; (1.70)}$ & $\mathbf{291620 \; (21842)}$ & $\mathbf{90.99 \; (1.20)}$ \\
        \bottomrule
    \end{tabular}}
\end{adjustbox}
\end{center}
\caption{Quantitative comparison of kernel size, number of steps, channel size and hidden dimensions based on SSIM, negative Jacobian determinants and Dice.}%
\label{tab:ablations}%
\end{table}

The results of our ablation studies shows that hyperparameter variations did not yield a significant effect on the model's overall performance. The modifications made to the NCA-Morph kernel size, channel size, and hidden size seemed to have an impact that could be regarded as random, given the absence of a significant difference. It is important to mention, that by increasing the number of steps, the perceptive range is increased leading to worse registration performance as shown in Table \ref{tab:ablations}.
With this insight in mind, we proceeded to perform our experiments with the NCA-Morph using 3x3 kernel with 5 steps and a 7x7 kernel with 10 steps, both having 32 channels and a hidden size of 128.

\subsection{NCA-Morph produces stable flows}
\label{ssec:nqm}
To assess the variance level of our NCA-Morph, we conducted ten consecutive inference registrations on three random images from the OASIS registration task.

\begin{figure}[htb]
\begin{center}
    \includegraphics[width=0.95\linewidth]{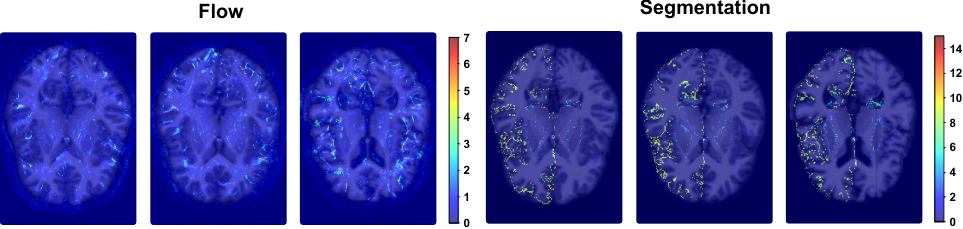}
\end{center}
    \caption{Stability assessment over 10 predictions on different test samples from the OASIS registration task. Left: variance for the flow, right: variance for the corresponding segmentation masks -- using NCA-Morph$_{7\times7}^{10}$.}
    \label{fig:nqm}
\end{figure}
We computed the standard deviation of the ten resulting flow fields, offering a measure of variability in the spatial transformations generated during the registration process. Additionally, we assessed the standard deviation of the corresponding segmentation masks, shedding light on the uncertainty associated with our model's performance. Figure \ref{fig:nqm} clearly indicates the reliability and consistency of our NCA-Morph in handling registration tasks.

\subsection{NCA-Morph runs on a Raspberry Pi}
\label{sec:rpi}
Incorporating DL models into resource-limited platforms like the Raspberry Pi (RPi) has emerged as an important endeavor. We showcase the unique benefits of our lightweight NCA-Morph architecture by deploying it on a Raspberry Pi 4, Model B (2 GB RAM), and illustrate its potential for edge computing applications. We conducted experiments to evaluate the performance and efficiency of our architecture in this constrained environment.

\begin{figure}[htb]
\begin{center}
    \includegraphics[width=0.9\linewidth]{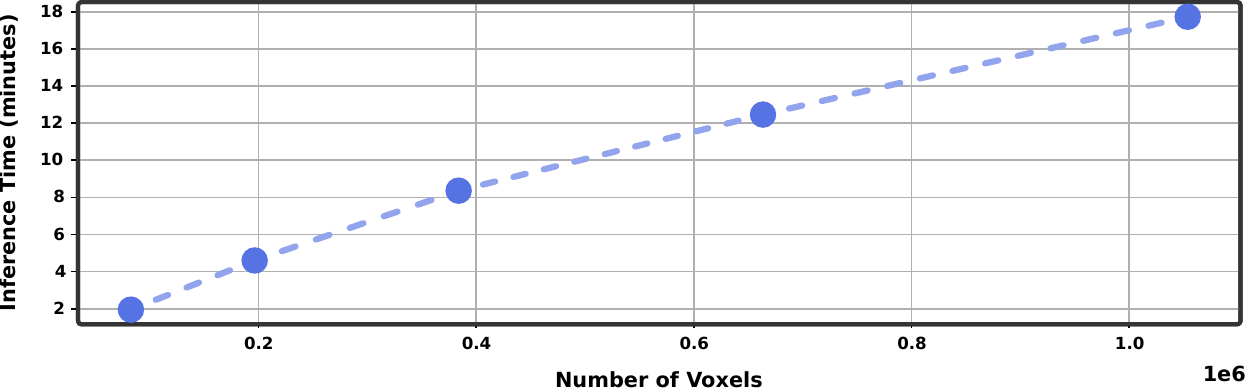} 
\end{center}
    \caption{Inference times of our NCA-Morph$_{7\times7}^{10}$ architecture across different input sizes on a Raspberry Pi.}
    \label{fig:rpi_fig}
\end{figure}

NCA-Morph's efficiency in resource utilization sets it apart as a particularly well-suited DL model for edge computing. Engineered with a focus on minimal memory and computational requirements, it seamlessly adapts to the RPi's constrained hardware of 2GB. This efficiency ensures not only smooth operation but also the potential for additional software components, making NCA-Morph a valuable addition to the Raspberry Pi ecosystem. With its focus on resource efficiency, real-time capabilities, and versatility, our NCA-Morph is positioned to contribute to advancements in various domains, while ensuring accessibility to a wide range of users.

\section{Conclusion}
We introduce \emph{NCA-Morph}, a robust medical image registration technique that uses an NCA architecture for registration instead of U-Net-based strategies. We evaluate our approach across three different registration tasks and show that it achieves state-of-the-art performance while being lightweight and using a significantly lower amount of parameters. Based on our bio-inspired approach, NCA-Morph succesfully builds a complex network between its cells while learning local displacement rules in a DL fashion. Future work should explore the NCA-related limitations of VRAM and training times to further improve the applicability of our method.

\section{Reproducability}
The used dataset is publicly available under the corresponding citation. The code with all implementations is made publicly available under \url{https://github.com/MECLabTUDA/NCA-Morph}.

\section{Acknowledgements}
This work is partially supported by Norwegian Research Council project number 322600 Capsnetwork.

\bibliography{bibliography}

\clearpage

\setcounter{section}{0}

\section*{\centering Supplementary Material}

\section{Flow from CNN harms performance} 
\label{sec:flows}

Our default NCA-Morph generates flows directly from its channels, while an alternative involves using an additional CNN layer after the NCA. Different fire rate during inference as shown in Appendix \ref{app:fire} leads to different shifts in the deformation field during both training and inference. Table \ref{tab:flow} captures the quantitative results of the flow generation and fire rate variations.

\begin{table}[htp]%
\begin{center}
\begin{adjustbox}{max width=0.75\linewidth}
{\begin{tabular}{lccccc}\toprule
  Method & Fire rate & Flow & SSIM $\uparrow$ & $\left\lvert J_\phi \right\rvert \leq 0 \downarrow$ & Dice $\uparrow$ [\%] \\ \hline \hline
    \textit{VoxelMorph} & -- & CNN & $82.84 \; (1.36)$ & $254217 \; (20553)$ & $90.82 \; (1.03)$ \\
    \hline
    NCA$^{10}_{7\times7}$ & $25\%$ & \multirow{4}{*}{\rotatebox{45}{NCA}} & $81.67 \; (1.66)$ & $339712 \; (33034)$ & $91.65 \; (1.09)$ \\
    NCA$^{10}_{7\times7}$ & $50\%$ & & $\mathbf{82.70 \; (1.66)}$ & $257241 \; (24826)$ & $\mathbf{92.14 \; (1.09)}$ \\
    NCA$^{10}_{7\times7}$ & $75\%$ & & $79.25 \; (1.48)$ & $170418 \; (17172)$ & $88.55 \; (1.37)$ \\
    NCA$^{10}_{7\times7}$ & $100\%$ & & $66.65 \; (1.15)$ & $\mathbf{166 \; (158)}$ & $66.33 \; (2.59)$ \\
    \hline
    NCA$*^{10}_{7\times7}$ & $25\%$ & \multirow{4}{*}{\rotatebox{45}{CNN}} & $64.38 \; (1.10)$ & $93308 \; (7070)$ & $61.23 \; (2.61)$ \\
    NCA$*^{10}_{7\times7}$ & $50\%$ & & $\mathbf{82.05 \; (1.57)}$ & $336983 \; (23095)$ & $\mathbf{91.29 \; (1.18)}$ \\
    NCA$*^{10}_{7\times7}$ & $75\%$ & & $65.21 \; (1.16)$ & $6520 \; (921)$ & $62.15 \; (2.61)$ \\
    NCA$*^{10}_{7\times7}$ & $100\%$ & & $65.64 \; (1.16)$ & $\mathbf{0.14 \; (0.4)}$ & $63.12 \; (2.61)$ \\
    \bottomrule
\end{tabular}}
\end{adjustbox}
\end{center}
\caption{Comparison of the two main variants predicting the flow directly using the NCA or an additional CNN layer, with different fire rates during inference; for every set of experiments the best values are marked bold, whereas number channels is set to 32 and hidden size to 128.}%
\label{tab:flow}
\end{table}

Table \ref{tab:flow} demonstrates the superior performance of directly generating the deformation field from the NCA compared to using an additional CNN layer, Appendix Figure \ref{fig:flow}.

\section{Impact of fire rate on deformation field}
\label{app:fire}
\begin{figure}[htb]
\begin{center}
    \includegraphics[width=\textwidth]{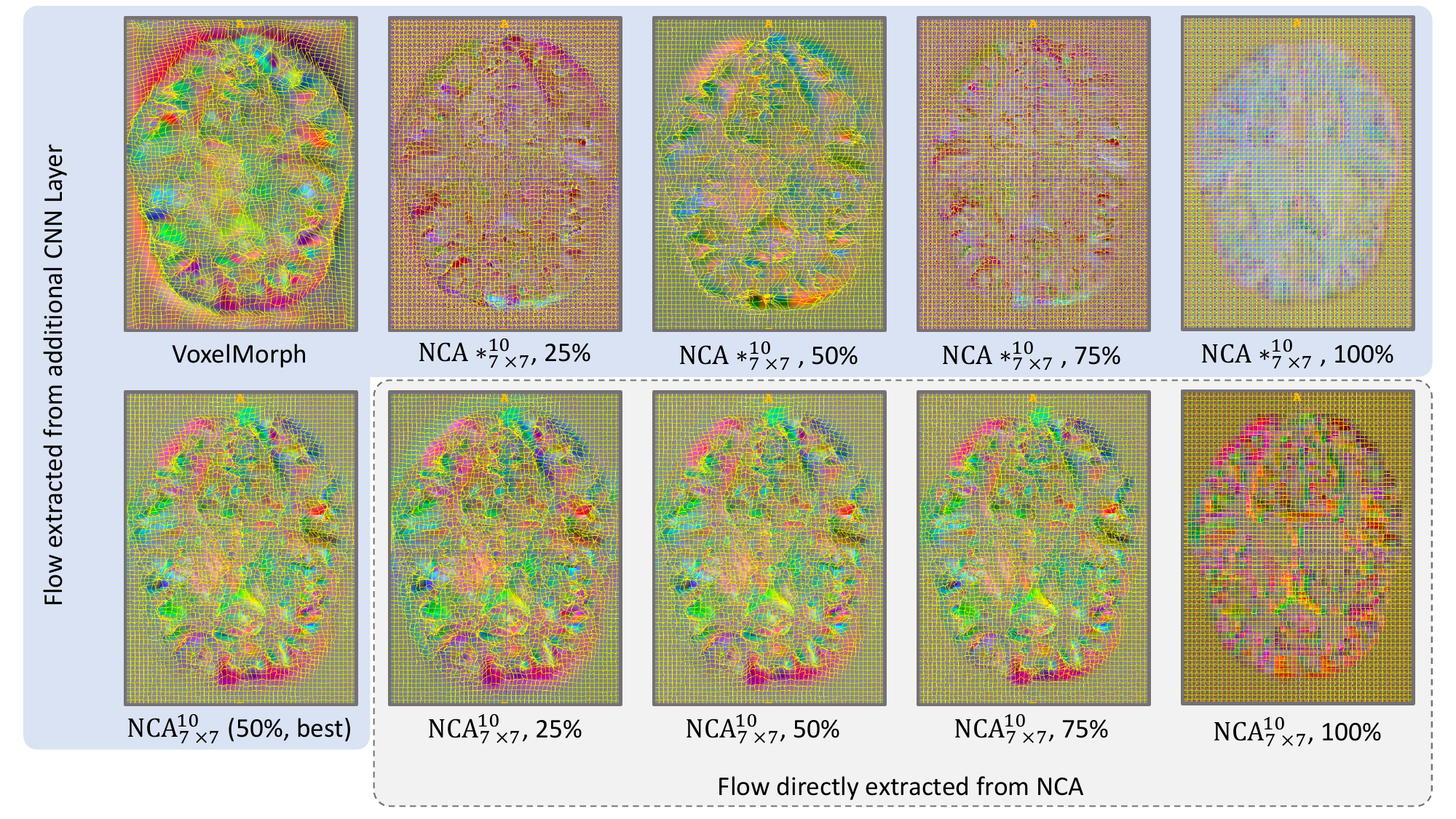}
\end{center}
    \caption{Qualititative analysis of deformation fields for different NCA-Morph variations showing the influence of fire rates on the flow adjustments; Notation: \textit{NCA-Morph$^\text{steps}_{\text{kernel}}$, \text{fire rate} \%}.}
    \label{fig:flow}
\end{figure}

Figure \ref{fig:flow} shows the deformation fields for different variations and visually confirms the results from Table \ref{tab:flow} of our main manuscript.
The best setup is the NCA-Morph using a $7 \times 7$ kernel size with $10$ steps, $32$ channels, a hidden dimension of $128$ using a fire rate of $50 \%$. NCA-Morph with an additional CNN layer to generate flows do not generate meaningful shifts anymore with increasing fire rates which harm the registration performance, Table \ref{tab:flow}). 

\section{More on Neural Cellular Automata (NCA)}
The adaptation of NCAs to a 3D registration problem was done in several steps. First, adapting the NCA from 2D to 3D is fairly straightforward, as the 2D convolution is replaced by a 3D convolution while the internal architecture remains the same. The more difficult problem is that 3D significantly increases the VRAM required for training, which requires further adjustment of the pipeline. We ensure this by performing a downsampling step before applying NCA, followed by another upscaling step. In addition, the pipeline requires careful adaptation of the underlying architecture, as it must be optimized to require as few parameters as possible. We achieve this by drastically simplifying the architecture, ultimately using only a single 3D convolution and very small linear layers. Lastly to enable the optimized architecture to perform registration it has to be integrated into the VoxelMorph training pipeline.

\clearpage

\end{document}